\documentclass[runningheads]{llncs}
\usepackage[T1]{fontenc}
\usepackage{graphicx}
\usepackage[utf8]{inputenc} 
\usepackage{hyperref}       
\usepackage{url}            
\usepackage{booktabs}       
\usepackage{amsfonts}       
\usepackage{nicefrac}       
\usepackage{microtype}      
\usepackage{xcolor}         

\usepackage{cite}
\usepackage{amsmath,amssymb,amsfonts}
\usepackage{algorithmic}
\usepackage{textcomp}
\usepackage{xcolor}

\usepackage{stmaryrd}

\newcommand{\enc}{\mathrm{enc}}
\newcommand{\dec}{\mathrm{dec}}

\newcommand{\softplus}{\mathrm{softplus}}

\newcommand{\dz}{d_\mathrm{z}}
\newcommand{\ds}{d_\mathrm{s}}
\newcommand{\dy}{d_\mathrm{y}}
\newcommand{\cond}[2]{\left(#1\;\middle|\;#2\right)}

\newcommand{\y}{\mathrm{y}}

\usepackage{multirow}
\usepackage[whole]{bxcjkjatype}

\begin{document}
\title{Virtual Human Generative Model: Masked Modeling Approach for Predicting Human Characteristics}
\titlerunning{Virtual Human Generative Model}
%
\author{
Kenta Oono\inst{1} \and
Nontawat Charoenphakdee\inst{1} \and
Kotatsu Bito\inst{2} \and
Zhengyan Gao\inst{1} \and
Hideyoshi Igata\inst{1} \and
Masashi Yoshikawa\inst{1} \and
Yoshiaki Ota\inst{1} \and
Hiroki Okui\inst{1} \and
Kei Akita\inst{1} \and
Shoichiro Yamaguchi\inst{1} \and
Yohei Sugawara\inst{1} \and
Shin-ichi Maeda\inst{1} \and
Kunihiko Miyoshi\inst{2} \and
Yuki Saito\inst{2} \and
Koki Tsuda\inst{2} \and
Hiroshi Maruyama\inst{1,2,3} \and
Kohei Hayashi\inst{1}
}

\institute{
Preferred Networks, Inc., Tokyo, Japan\\
\and
Kao Corporation, Tokyo, Japan\\
\and
The University of Tokyo, Tokyo, Japan
}

%
%
%

\maketitle              
\begin{abstract}
Virtual Human Generative Model (VHGM)~\cite{vhgmplos} is a generative model that approximates the joint probability over more than 2000 human healthcare-related attributes. 
This paper presents the core algorithm, VHGM-MAE, a masked autoencoder (MAE) tailored for handling high-dimensional, sparse healthcare data.
VHGM-MAE tackles four key technical challenges: (1) heterogeneity of healthcare data types, (2) probability distribution modeling, (3) systematic missingness in the training dataset arising from multiple data sources, and (4) the high-dimensional, small-$n$-large-$p$ problem.
To address these challenges, VHGM-MAE employs a likelihood-based approach to model distributions with heterogeneous types, a transformer-based MAE to capture complex dependencies among observed and missing attributes, and a novel training scheme that effectively leverages available samples with diverse missingness patterns to mitigate the small-$n$-large-$p$ problem.
Experimental results demonstrate that VHGM-MAE outperforms existing methods in both missing value imputation and synthetic data generation.

\keywords{Generative modeling \and Tabular data learning \and Missing value imputation \and Masked autoencoders. \and Healthcare applications}
\end{abstract}
\section{Introduction}
\begin{figure}[t]
  \includegraphics[width=1.0\textwidth]{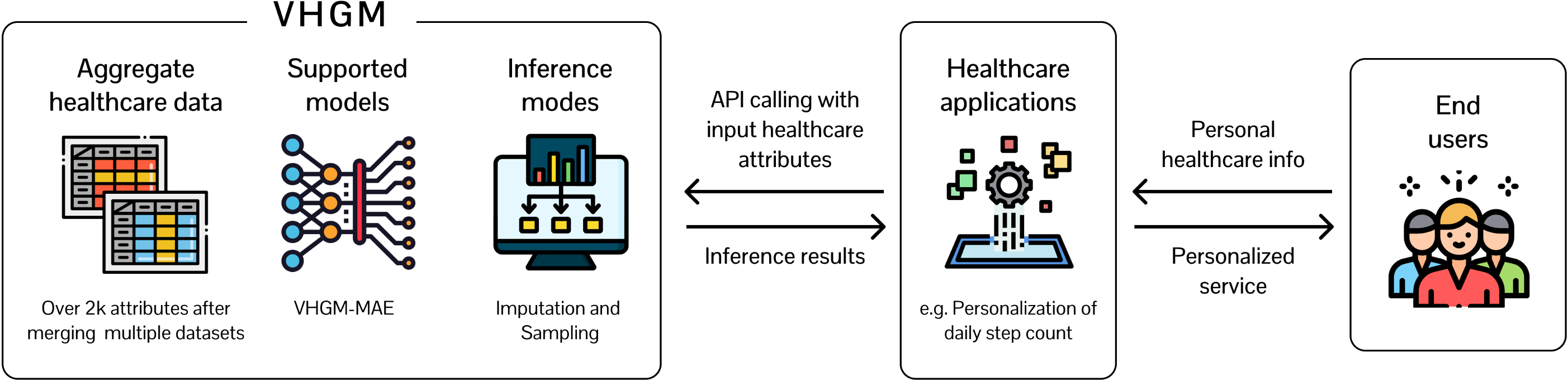}
  \centering
  \caption{Overview of VHGM's core use-case. 
Users provide healthcare data, which queries VHGM to obtain healthcare attribute inference and returns personalized services.
  }
  \label{fig:teaser}
\end{figure}

The state of human health at a time can be observed in many different ways, for example, by measuring blood pressure and answering a questionnaire on exercise habits.
Virtual Human Generative Model (VHGM)~\cite{vhgmplos} is a generative model that approximates the joint probability over more than 2000 observable values, hereafter referred to as \emph{attributes} in this paper.
VHGM is deployed as a commercial API service to enable estimating unknown attributes and generating synthetic data (see Fig.~\ref{fig:teaser}).


Our technical challenges in building such a model are fourfold.
The first challenge arises from the multi-modality of healthcare attributes. For instance, attribute types can be real, ordinal, categorical, or count, each requiring different modeling strategies.
The second challenge is accurate data distribution modeling, which requires not only point prediction but also capturing uncertainty and enabling sampling from interpretable probability distributions.
The third challenge concerns systematic missingness, which is common in real-world healthcare data.
Such patterns, often characterized as Missing Not at Random (MNAR), frequently arise when merging datasets with differing missingness mechanisms, making model training particularly challenging.
The fourth challenge is the \textit{small-n-large-p} problem. 
Healthcare datasets tend to be high-dimensional (i.e., large dimensionality $p$) but with relatively small sample size~$n$.
With small $n$, it can be challenging to build a reliable machine learning model.
With large $p$, the usage of iterative methods that do not scale well with $p$ such as multiple imputation by chained equations (MICE)~\cite{van2000multivariate} or HyperImpute~\cite{jarrett2022hyperimpute} can be prohibitive.


In this paper, we propose VHGM-MAE, a deep generative framework based on a masked autoencoder (MAE).
To handle heterogeneous attributes and distribution modeling, we incorporate likelihood-based modeling following the approach of~\cite{NAZABAL2020107501}, which provides a principled way to model attributes of different data types and their associated uncertainty.
To effectively address systematic missingness, VHGM-MAE employs MAE~\cite{He_2022_CVPR,vaswani2017attention}, which is built upon the transformer architecture and can be trained using the masked modeling paradigm~\cite{devlin-etal-2019-bert}, where portions of the input are intentionally masked and the model learns to reconstruct the missing parts.
This training strategy has shown to be effective in natural language processing and has been successfully extended to image recognition~\cite{He_2022_CVPR} and tabular learning~\cite{Arik_Pfister_2021}, demonstrating its versatility across diverse data modalities.
Together, these design choices make VHGM-MAE a powerful model for virtual human modeling, capable of representing diverse, incomplete, and high-dimensional healthcare data.

\section{Related Work}
\textbf{Tabular Data Missing Value Imputation}
There are three key assumptions of missing data: (1) Missing Completely at Random (MCAR), (2) Missing at Random (MAR), and (3) Missing Not at Random (MNAR)~\cite{rubin1976inference,van2018flexible}.
Advanced imputation techniques fall into two categories~\cite{jarrett2022hyperimpute}: iterative methods and deep generative models. 
The iterative approach models the conditional distribution of each attribute using a combination of observed and imputed values. 
Representative methods in this category is Multiple Imputation by Chained Equations (MICE)~\cite{van2000multivariate}, MissForest~\cite{stekhoven2012missforest}, and HyperImpute~\cite{jarrett2022hyperimpute}.
In our target scenario where $p > 2000$, such methods can become computationally prohibitive (see Chapter 9.1 of~\cite{van2018flexible} for more discussion). 
Unlike the iterative approach, deep generative model leverages deep learning and scale more efficiently as the number of attributes $p$, increases. This makes them particularly well-suited for high-dimensional datasets.
Several architectures have been proposed to address this challenge such as  Heterogeneous Incomplete Variational Autoencoder (HIVAE)~\cite{NAZABAL2020107501}, MIDA~\cite{gondara2018mida}, MIWAE~\cite{mattei2019miwae}, diffusion models~\cite{tashiro2021csdi,zheng2022diffusion}, and GANs~\cite{yoon2018gain}. 
Recently, Remasker~\cite{mae_du2023remasker}, a transformer-based architecture, has been introduced, showing strong performance in missing value imputation. 
Nevertheless, Remasker only supports point estimation and does not explicitly support heterogeneous attributes.

\textbf{Deep Generative Modeling for Tabular Data} 
Variational Autoencoder (VAEs)~\cite{Kingma2014,akrami2022robust} are well-suited for continuous attributes and have been extended to handle more complex data types, offering flexibility in both data generation and imputation~\cite{NAZABAL2020107501}.
Generative adversarial approach which employs adversarial training framework between generator and discriminator has also been studied for tabular data~\cite{gan_generation_1,gan_generation_2,gan_generation_3}.
Diffusion model approach has recently emerged as a compelling alternative to VAEs and GANs, which allows more stable training than GANs and better control of sampling diversity. This approach has also been used in tabular data generation~\cite{diffusion_generation_1,diffusion_generation_2,diffusion_generation_3,diffusion_generation_4}.
Furthermore, transformers~\cite{vaswani2017attention}, which are based on an encoder-decoder framework, have shown promise for tabular data modeling~\cite{Arik_Pfister_2021,deng2020turl,gorishniy2021revisiting,huang2020tabtransformer,iida2021tabbie,kossen2021selfattention,mae_gulati2024tabmt}. 
With attention mechanism, transformers can model complex attribute interactions in a flexible and scalable manner.
However, the use of the transformer architecture to explicitly model distributions and support heterogeneous attributes in tabular data have been relatively underexplored.

To the best of our knowledge, few studies have investigated deep generative modeling frameworks capable of representing missing value distributions while handling heterogeneous attribute types. Among them, HIVAE~\cite{NAZABAL2020107501}, a variant of the variational autoencoder (VAE) meets these requirements. 
In this paper, we propose VHGM-MAE: a scalable and expressive generative model by integrating the likelihood modeling principle of HIVAE with the MAE architecture trained via masked modeling, achieving superior performance over prior methods in both data imputation and generation tasks.

\section{Proposed Method: VHGM-MAE}
\subsection{Model Architecture}
\begin{figure}[t]
  \centering
  \includegraphics[width=1.0\linewidth]{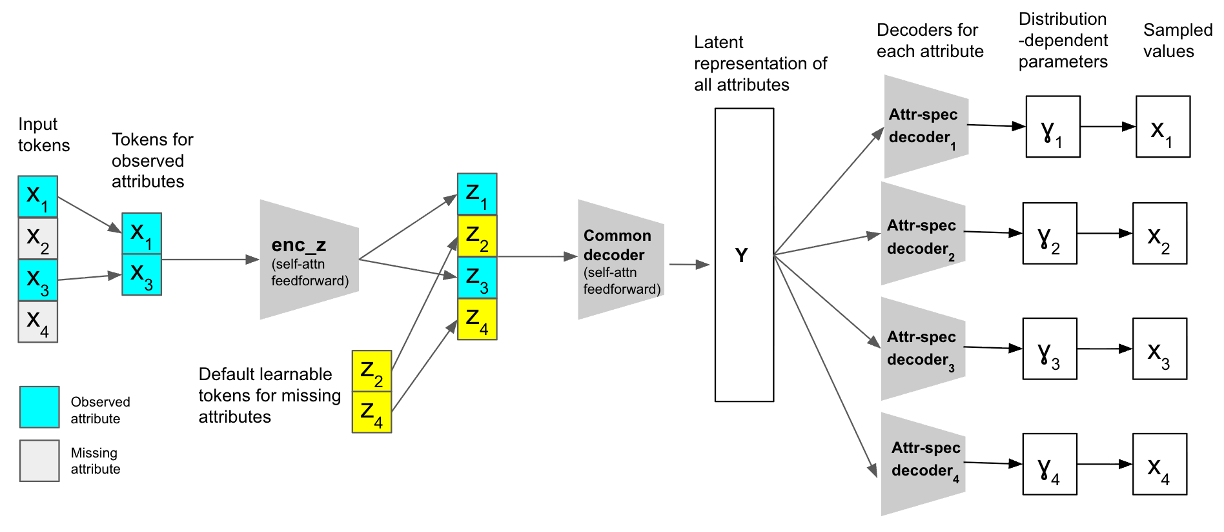}
  \caption{Overview of VHGM-MAE where $p=4$, $x_2$ and $x_4$ are missing inputs. 
  Transformer is used for the encoder and common decoder, while linear models are used for the attribute-specific (Attr-spec) decoders to scale well when $p$ is large.}
  \label{fig:vhgm-mae-arch}
\end{figure}

VHGM-MAE employs transformers~\cite{vaswani2017attention}, which primarily consists of self-attention mechanisms and feedforward neural networks. 
Figure~\ref{fig:vhgm-mae-arch} provides an overview of VHGM-MAE architecture.
The observed attributes are passed to the transformer encoder, while missing values are replaced with the ``default'' learnable mask tokens.
Learnable mask tokens are randomly initialized for each attribute and are optimized during training.
Encoded latent representations, along with the mask tokens, are then passed to the transformer common decoder, followed by an attribute-specific decoder, to model the attribute likelihood distribution.

Mathematically, we define the deterministic encoder $\enc^{\mathrm{MAE}}_\phi: \mathcal{X} \to \mathbb{R}^{\dz}$, where $\dz$ represent the dimension of latent attribute $z$ and $\phi$ are encoder parameters.
The encoder's output with dimension $\dz$ is constructed from the concatenation of the encoded inputs and the learnable mask tokens for each attribute, where the latter in fact do not pass the encoder but we combine both vectors into encoder's output for notational simplicity.

We also define the decoder $\dec^{\mathrm{MAE}}_\theta(z)=(\gamma_1, \ldots, \gamma_p)$, where $\theta$ are decoder parameters and $\gamma_j$ is distribution parameter for attribute~$j$.
For the decoder $\dec^{\mathrm{MAE}}_\theta$, to effectively handle high-dimensional data,
it is constructed from the composition of the common decoder $\dec^{\mathrm{MAE}}_{\theta, \y}: \mathbb{R}^{\dz} \to \mathbb{R}^{\dy}$ and the attribute-specific decoder $\dec^{\mathrm{MAE}}_{\theta, j}: \mathbb{R}^{\dy} \to \Gamma_j$. 
Given an input $x \in \mathcal{X}$, the encoder and decoder are processed as follows to compute the distribution parameter for attribute $j$: $z = \enc^{\mathrm{MAE}}_\phi(x), y = \dec^{\mathrm{MAE}}_{\theta,y}(z), \gamma_j = \dec^{\mathrm{MAE}}_{\theta,j}(y)$.

In summary, VHGM-MAE architecture consists of one common encoder, one common decoder, and $p$ attribute-specific decoders. 
It is crucial that the attribute-specific decoders remain lightweight to avoid excessive memory usage while a common decoder can be more complex. 

To model the likelihood to support heterogeneous attributes, we borrow the idea of HIVAE~\cite{NAZABAL2020107501}. Let~$\Gamma_j$ be the parameter space for the $j$-th attribute, differing by the attribute type:
\begin{equation}
    \gamma_{j} =
    \begin{cases}
        (\mu_j, \sigma^2_j) \in \mathbb{R} \times \mathbb{R}_{+} & \text{(real)}, \\
        \lambda_j \in \mathbb{R}_{+} & \text{(count)}, \\
        (\mu_j, \sigma^2_j) \in \mathbb{R} \times \mathbb{R}_{+} & \text{(positive)}, \\
        \pi_{j} \in \mathcal{P}^{c_j} & \text{(categorical)}, \\
        (r_j, h_j) \in \mathbb{R}^{c_j - 1} \times \mathbb{R}& \text{(ordinal)}, \\
    \end{cases} \label{eq:decoder}\\
\end{equation}
where $c_j$ is the number of categories of the $j$-th attribute.
We add the softmax function as the final layer of the decoder when the $j$-th attribute is categorical.
For ordinal attributes, we convert the parameters $r_j = (r_{j1}, \ldots, r_{j(c_j-1)})$ and $h_j$ to an increasing sequence $r'_j=(r'_{j1}, \ldots, r'_{j(c_{j-1})})\in \mathcal{O}^{c_{j-1}}$ by 
    $r'_{jk} = \sum_{l=1}^k \softplus(r_{jl}) - h_j$
for $k = 1, \ldots, c_{j-1}$.
The probability distribution $p_{\theta}\cond{x}{z} = \prod_{j=1}^p p_{\theta, j}\cond{x_j}{z}$ is modeled using $\gamma_{j}$'s as follows: we use Gaussian likelihood for real attributes, Poisson likelihood for count attributes, log-normal likelihood for positive attributes, multinomial logit for categorical attributes, and ordinal logit with thermometer encoding for ordinal attributes.


When calculating self-attention in the decoder, we use a non-diagonal decoder, similarly to the idea of cross-attention~\cite{crossattention}, which has been shown that self-attention between mask patches is not necessary for good performance in the computer vision.
It modifies the attention mechanism to selectively compute interactions between the pairs of observed attributes and missing attributes while omitting the computation of interactions between observed-observed attributes and missing-missing attributes. 
This design ensures that the decoder prioritizes relevant attribute relationships while reducing unnecessary computations.

\subsection{Training method}
\subsubsection{Loss Function}
The key idea is to minimize the negative log-likelihood of missing attributes.
Given the dataset $\mathcal{D} = (x_i)_{i=1}^n$ where $x_i = (x_{ij})_{j=1}^{p}$ is the $i$-th training instance, and $p_\theta\cond{x_{ij}}{z_{i}}=\dec^{\mathrm{MAE}}_{\theta,j}(y_j)$ , we train the model to minimize the following loss function $L^{(t)}$ at the $t$-th epoch: $  L^{(t)}(\theta, \phi)
        = \sum_{i=1}^{n} \sum_{j=1}^{p} -m^{(t)}_{ij} \log p_\theta\cond{x_{ij}}{z_{i}}$.
Here, $m^{(t)}_{ij}$ equals $1$ when the $j$-th element of the $i$-th instance is masked at the $t$-th iteration and $0$ otherwise. 
It is worth noting that unlike HIVAE, where loss is calculated on the observed attribute prediction, we calculate from the missing attribute prediction.

\subsubsection{Mask augmentation}
To enhance generalization to unseen missing patterns, we introduce a technique called mask augmentation.
At each training epoch, we randomly select a subset of attributes that are not missing in the records of each minibatch and mark them as missing.
This process effectively increases the diversity of missing patterns encountered during training, thereby improving the model's generalization.

\subsubsection{Two-stage training}
For the case where we have several datasets from different source (as in Sec.~\ref{sec:real-dataset}), we employ a two-stage training procedure. 
The first stage focuses on intra-dataset interactions using many epochs, while the second stage, with fewer epochs, addresses both intra- and inter-dataset interactions.
In the first stage, we perform dataset-wise training, processing each dataset sequentially within one epoch using gradient accumulation. 
This method constructs smaller attention matrices for each dataset, reducing computational costs and emphasizing intra-dataset interactions. 
While this approach accelerates training and can focus intra-dataset relationships, it may not fully capture interactions between datasets. 
Thus, a second stage is necessary to refine the model's effectiveness across heterogeneous datasets.
In the second stage, we combine all datasets and train for a shorter number of epochs compared to the first stage as it is more time-consuming epoch-wise. 
This stage enables the model to learn both intra- and inter-dataset interactions.
In fact, we found that training by using only second stage for a large number of epochs (e.g., 300) is sufficient to achieve good performance, but it will be much more time consuming.

\subsection{Sampling}
The sampling can be done from the generative model $p_\theta(x|z) = \dec^{\mathrm{MAE}}_\theta(z) = (\gamma_1, \ldots, \gamma_p)$, which is called predictive-distribution sampling. 
Predictive-distribution sampling draws samples from the distribution parameterized by the output $\gamma_i$ of the model (e.g., the Gaussian distribution for real attributes) in Eq.~\eqref{eq:decoder}.
The variability of the predictive-distribution sampling represents the uncertainty of the generative model. 
One can also conduct sampling in an autoregressive way similarly to language models~\cite{mae_gulati2024tabmt}, i.e., sampling attributes iteratively, where the output of the previous iteration is used as the input of the next iteration. 

\section{Experiments}

\begin{table*}[t]
\centering
\scriptsize
\caption{RMSE on benchmark datasets with 70\% MNAR missingness. Bold indicates the lowest RMSE methods while underlined text indicates the second lowest ones.}
\begin{tabular}{lcccccccccc}
\toprule
 & Constant & \multicolumn{3}{c}{Iterative} & \multicolumn{4}{c}{Deep learning} &  Proposed \\
\cmidrule(lr){2-2} \cmidrule(lr){3-5} \cmidrule(lr){6-9} \cmidrule(lr){10-10}
Dataset & Mean & MICE & MissForest & HyperImpute & MIWAE & GAIN & HIVAE & Remasker & VHGM-MAE \\
\midrule
climate     & 0.305 (.01) & 0.394 (.02) & 0.313 (.01) & 0.315 (.01) & 0.429 (.03) & 0.361 (.01) & 0.294 (.02) & \underline{0.289 (.02)} & \textbf{0.287 (.02)} \\
compression & 0.237 (.00) & 0.324 (.01) & \underline{0.220 (.00)} & 0.226 (.01) & 0.261 (.00) & 0.319 (.01) & 0.237 (.00) & 0.245 (.01) & \textbf{0.215 (.00)} \\
wine        & 0.160 (.01) & 0.196 (.01) & 0.143 (.01) & \underline{0.141 (.01)} & 0.171 (.01) & 0.195 (.03) & 0.162 (.01) & 0.151 (.01) & \textbf{0.139 (.01)} \\
yacht       & \underline{0.303 (.00)} & 0.416 (.01) & 0.333 (.01) & 0.319 (.01) & 0.641 (.01) & 0.464 (.03) & 0.326 (.03) & 0.345 (.01) & \textbf{0.301 (.01)} \\
spam        & 0.069 (.02) & 0.083 (.03) & \underline{0.057 (.02)} & 0.060 (.01) & 0.071 (.03) & 0.069 (.02) & 0.079 (.02) & \textbf{0.053 (.01)} & 0.062 (.02) \\
credit      & 0.310 (.04) & 0.413 (.05) & \underline{0.297 (.05)} & 0.300 (.05) & 0.385 (.03) & 0.389 (.03) & 0.341 (.07) & \textbf{0.277 (.04)} & 0.329 (.06) \\
raisin      & 0.183 (.00) & 0.196 (.00) & \underline{0.140 (.00)} & 0.142 (.00) & 0.249 (.00) & 0.377 (.05) & 0.182 (.00) & 0.195 (.02) & \textbf{0.138 (.00)} \\
california  & 0.151 (.00) & 0.191 (.00) & \underline{0.150 (.01)} & 0.159 (.01) & 0.160 (.00) & 0.283 (.01) & 0.151 (.01) & 0.157 (.01) & \textbf{0.134 (.00)} \\
\bottomrule
\end{tabular}
\label{tab:benchmark}
\end{table*}

\subsection{Benchmark Experiments}

\subsubsection{Setup}\label{sec:exp-methods}
For the baseline methods, we categorize them into three groups: constant, iterative, and deep learning baselines.
For the constant baseline, we used the mean values of each column, as all columns contain real-valued attributes.
For the iterative baselines, we employed MICE (with Bayesian ridge regression)~\cite{van2000multivariate}, MissForest~\cite{stekhoven2012missforest}, and HyperImpute~\cite{jarrett2022hyperimpute}.
For the deep learning baselines, we utilized MIWAE~\cite{mattei2019miwae}, GAIN~\cite{yoon2018gain}, HIVAE~\cite{NAZABAL2020107501}, and Remasker~\cite{mae_du2023remasker}.
All methods, except HIVAE, provide only point estimates and cannot quantify uncertainty without additional modifications. Remasker can be considered a state-of-the-art baseline among the evaluated methods.
Details of the hyperparameter configurations are provided in Appendix~\ref{sec:hyperparam}.

For benchmark datasets, we used eight UCI datasets from the Remasker repository~\cite{mae_du2023remasker}, which are \textit{climate}, \textit{compression}, \textit{wine}, \textit{yacht}, \textit{spam}, \textit{credit}, \textit{raisin}, and \textit{california} (See dataset descriptions in Table 6 of~\cite{mae_du2023remasker}). 
To simulate the high missing rate regime, which is common in the small-$n$-large-$p$ problem, we used MNAR missing patterns with a 70\% missing rate for the evaluation. At such a high missing rate, a simple mean imputer can be competitive in some datasets.

For an evaluation metric, we used root mean squared error (RMSE): \begin{align*}
    \sqrt{\frac{1}{n_\mathrm{test}}\sum_{i=1}^{n_\mathrm{test}} (y^{\mathrm{pred}}_i - y^{\mathrm{test}}_i)^2},
\end{align*} as an evaluation metric following the previous work~\cite{mae_du2023remasker}. 

\subsubsection{Results}
Table~\ref{tab:benchmark} reports the RMSEs for each method. 
VHGM-MAE performs competitively, achieving the best results on six out of the eight datasets.
Among the baselines, MissForest, Remasker, and HyperImpute are also observed to be strong performers compared to the others.
Note that these existing imputation methods work only on numeric (\emph{real}) data or binary attributes and offer only deterministic imputation without estimating the probability distribution.
VHGM-MAE is competitive for point imputation of \emph{real} data while also being able to handle other data types with probabilistic modeling.

\begin{table}[t]
    \caption{Dataset statistics for small-$n$ large-$p$ experiments~\cite{vhgmplos}.  
    We first allocate the records for validation and test splits and then conduct sampling with replacement from the remaining part of the records. 
    }
    \centering
    \begin{tabular}{crrrrr}
        \toprule
        Dataset  & Attributes ($p$) &Records ($n$) &  Train  & Validation & Test \\
        \midrule
        A & 1,840 & 897  & 18,000 & 100 & 100\\
        B & 257 & 1,121,227  & 100,000 & 10,000 & 10,000 \\
        C-1 & 61 & 10,483  & 18,000 & 1,000 & 1,000 \\ 
        C-2 & 162 & 1,584  & 18,000 & 500 & 500\\ 
        \bottomrule
    \end{tabular}
    \label{tab:dataset-statistics}
\end{table}

\subsection{Real-world Dataset: Small-n Large-p Experiments}
\label{sec:real-dataset}
\subsubsection{Setup}
In this setting, we used a tabular dataset aggregated from four different sources, each exhibiting distinct characteristics in terms of the number of samples ($n$) and attributes ($p$).
Table~\ref{tab:dataset-statistics} summarizes the statistics of these datasets, while detailed information on the data collection process is provided in~\cite{vhgmplos}.
Several attributes are shared across datasets, whereas others are unique to specific sources.
The aggregated dataset comprises a total of 2,110 attributes, including 509 categorical, 199 count, 244 ordinal, 16 positive, and 1,142 real-valued attributes.
Given the large number of attributes, we expect an extremely high missing rate during testing—approximately 98–99\%, meaning that a typical input instance contains only about 10–20 observed attributes.

Merging datasets offers the advantage of a larger sample size compared to any individual dataset.
However, it also introduces a systematic missingness pattern.
Even if the original datasets contain no missing values, any dataset lacking attributes present in others will be treated as having missing entries.
For instance, Dataset B contains only 257 attributes; thus, each of its records will have missing values for 2,110 − 257 = 1,853 attributes.
Consequently, the missingness in this problem can be characterized as Missing Not At Random (MNAR), a particularly challenging setting to model~\cite{mnar_ma2021identifiable,mnar_pereira2024imputation}.

We evaluate performance of our proposed method through two experiments:
(1) missing value imputation, which assesses the quality of point estimation, and
(2) machine learning efficiency, which evaluates the sampling quality for downstream predictive tasks.

\subsubsection{Imputation Experiments}
\begin{table}[t]
    \caption{Imputation error across different types on real-world dataset (2110 attributes).
    }
    \centering
    \begin{tabular}{ccccccc}
        \toprule
        Method  & Cat. & Count & Ordinal & Positive & Real & Total \\
        \midrule
        Mode-mean & 0.2832 & 0.0403 & 0.1179 & 0.1058 & 0.1700 & 0.1551 \\
        XGBoost & 0.2126 & 0.0395 & 0.1182 & 0.2177 & 0.1540 & 0.1430 \\
        HIVAE & 0.2405 & 0.0396 & 0.1205 & 0.1086 & 0.1695 & 0.1521 \\
        \midrule
        VHGM-MAE (w/o mask aug.) & 0.2200  & 0.0378       & 0.1260           & 0.1090       & 0.1558        & 0.1451         \\    
        VHGM-MAE & \textbf{0.1890} & \textbf{0.0372} &\textbf{0.1142} & \textbf{0.1012}& \textbf{0.1431} & \textbf{0.1324} \\
        \bottomrule
    \end{tabular}
    \label{tab:comparison-with-mode-imputer}
\end{table}
For baselines, we compared the model with the mode-mean imputer, which fills missing values with the mode for categorical attributes and mean values for continuous attributes and rounded mean values for the count and ordinal attributes. 
We also used XGBoost and HIVAE~\cite{NAZABAL2020107501} as reference baselines.
We tried MICE~\cite{van2000multivariate} and HyperImpute~\cite{jarrett2022hyperimpute} but found that the model failed to converge, which is expected as this approach does not scale well with the number of attributes~\cite{van2018flexible}. As a result, we exclude the iterative baselines in this experiment.
We provide a smaller-scale experiment ($p=260$) to compare with MICE in Appendix~\ref{sec:app-small-exp}.
Furthermore, we provide the result of VHGM-MAE without mask augmentation as an ablation study of the effectiveness of mask augmentation.

For evaluation metrics, We used average prediction error for categorical attributes: $\frac{1}{n_\mathrm{test}}\sum_{i=1}^{n_\mathrm{test}}  \llbracket y^{\mathrm{pred}}_i \not = y^{\mathrm{test}}_i \rrbracket $, mean absolute error normalized by the number of classes for ordinal attributes $\frac{1}{n_\mathrm{test}}\sum_{i=1}^{n_\mathrm{test}} \frac{| y^{\mathrm{pred}}_i - y^{\mathrm{test}}_i |}{c}$,  RMSE normalized by the difference between maximum and minimum of the ground truths for real and positive attributes: $\frac{\sqrt{\frac{1}{n_\mathrm{test}}\sum_{i=1}^{n_\mathrm{test}} (y^{\mathrm{pred}}_i - y^{\mathrm{test}}_i)^2}}{\max(Y^{\mathrm{test}}) - \min(Y^{\mathrm{test}})}$.

Table~\ref{tab:comparison-with-mode-imputer} shows the mean performance score of five independent trials.
It can be observed that our VHGM-MAE achieves better performance than the baselines.
While the original HIVAE underperforms XGBoost in prediction, our proposed VHGM-MAE with mask augmentation surpasses XGBoost across all attribute types. 
It can also be observed that mask augmentation improves performance of VHGM-MAE across all attributes.

\subsubsection{Machine Learning Efficiency Experiments}\label{sec:machine-learning-efficiency}

\begin{table}[t]
    \caption{ML efficiency: depression scale regression. Combining VHGM-generated data with original data achieves the best performance.}
    \centering
    \begin{tabular}{ccccc}
        \toprule
        Dataset & Individual (R2) & Combined with Original (R2)\\
        \midrule
       Original & 0.531  & -  \\
       \midrule
       TVAE & 0.225 & 0.525 \\
       CTGAN & -0.130 & 0.506 \\
       Gaussian Copula & 0.476 & 0.531 \\
       VHGM-MAE & \textbf{0.501} & \textbf{0.553} \\
        \bottomrule
    \end{tabular}
    \label{tab:ml-eff-depression-scale}
\end{table}

We conducted a machine learning efficiency task to evaluate sample generation quality.
The task is depression scale regression from 65 attributes related to exercise, nutrients, sleeping habits, stress and tiredness.
We generated 10,000 rows of synthetic data with 66 and 16 attributes (inputs and one output) for the first and second task respectively. 
We compared between VHGM-MAE and baselines in synthetic data vault~\cite{patki2016synthetic}, where we used TVAE, CTGAN, and Gaussian Copula.
For regression, we trained a linear regression model from each synthetic dataset and use a test R2 score for the evaluation.
We evaluate the performance of using only the original dataset or the synthetic dataset (individual) and using both the original dataset and synthetic dataset (Combined with Original).

Table~\ref{tab:ml-eff-depression-scale} shows the performance of depression scale regression.
VHGM-MAE outperforms the baselines in both cases.
Although using VHGM-MAE synthetic data is worse than using the original dataset, combining VHGM-MAE data and original dataset can further boost performance beyond simply using original dataset.
We hypothesize the effectiveness of VHGM-MAE data augmentation could come from (1) VHGM-MAE succeeds in capturing attribute relationships that are useful for the tasks and (2) data augmentation can reduce overfitting.

\section{Conclusion}
We proposed VHGM-MAE, a transformer-based generative model for joint distribution modeling of observable healthcare data, lifestyle factors, and personality traits.
Experiments revealed the effectiveness of VHGM-MAE on benchmark datasets and real-world dataset in missing value imputation and machine learning efficiency.
We believe the versatility of VHGM-MAE opens the door to realize a wide range of healthcare applications, thereby contributing to the social good by improving quality of life.
The current limitations of VHGM-MAE are that it captures only statistical interactions among attributes, which should not be interpreted as causal relationships without prior knowledge, and that it does not yet support time-series data.

%
%
%

%
%
%
\bibliographystyle{splncs04}
\bibliography{reference}

\appendix
\section{Hyperparameters}\label{sec:hyperparam}

\begin{table}
\centering
\caption{Model architecture hyperparameters for VHGM-MAE.}
\label{tab:model_architecture}
\begin{tabular}{lcc}
\toprule
\textbf{Component} & \textbf{Benchmark datasets} & \textbf{\S\ref{sec:real-dataset} dataset} \\
\midrule
\textbf{Enc.} & 1 hid. layer, 320 nodes & 2 hid. layer, 384 nodes \\
\textbf{Common dec.} & 1 hid. layer, 832 nodes & 2 hid. layer, 384 nodes \\
\textbf{Specific dec.} & 3 hid. layers, 100 nodes & Linear model \\
\textbf{Activation} & ReLU & ReLU \\
\textbf{Attention} & 4 heads, 32 dim/head & 4 heads, 24 dim/head \\
\textbf{Mask aug. rate} & 10\% & 99\% \\
\textbf{Batch size} & 16 & 32 \\
\bottomrule
\end{tabular}
\end{table}

Table~\ref{tab:model_architecture} shows hyperparameters for VHGM-MAE architecture that differs between benchmark datasets and aggregated dataset discussed in Section~\ref{sec:real-dataset}.
For optimization, we used AdamW, where the learning rate was set to $5 \times 10^{-4}$ with weight decay parameter as $2.5 \times 10^{-4}$ and beta parameter for AdamW as 0.9.
For benchmark datasets, where there are small attributes, we only use full-dataset training with 200 epochs.
For the healthcare dataset in Sec.~\ref{sec:real-dataset}, the number of epochs was set to 300 for the first stage of dataset-wise training and 10 for the second stage of full-dataset training.
For HIVAE, We used 2-hidden layer with 850 hidden nodes for each layer for common decoder, attribute-specific decoder and encoder.
However, when using large dataset described in Section~\ref{sec:real-dataset}, we used a linear model for attribute-specific decoder to make the computation feasible similarly to VHGM-MAE.
Dimensions of the latent attributes $s$ and $z$ for HIVAE ($\ds$ and $\dz$) were set to 83 and 57, respectively. 
For optimization, we used AdamW, where the learning rate was set to $4.6 \times 10^{-5}$ with weight decay parameter as $0.097$ and beta parameter for AdamW as 0.9.
For benchmark datasets, the batch size, number of epochs are set the same way as VHGM-MAE.
For dataset in Section~\ref{sec:real-dataset}, the batch size was set to 1024, and the number of epochs was set to 1000, where we employed early stopping with patience equal to 50.

\section{260-attribute experiment: comparison with MICE}
\label{sec:app-small-exp}
To compare with MICE, we ran algorithm on only datasets B and C-1 containing 260 attributes, which is a subset of the real-world experiment setting in Section~\ref{sec:real-dataset}. 
We used MICE with random forest (MICE-RF), lasso regression (MICE-L), and ridge regression (MICE-R) as baselines.
Table~\ref{tab:comparison-small-data-transposed} shows that MICE has reasonable performance although XGBoost outperformed MICE with random forest and MICE with lasso, while our VHGM-MAE still outperformed MICE in this setting.


\begin{table}[tb]
    \caption{Proposed method can outperform MICE approach. Imputation error of 260 attributes (trained on Datasets B and C-1). }
    \centering
    \begin{tabular}{lcccccc}
        \toprule
        & Mode-Mean & XGBoost & MICE-RF & MICE-L & MICE-R & VHGM-MAE \\
        \midrule
        Total error & 0.1274 & 0.1134 & 0.1138 & 0.1139 & 0.1109 & \textbf{0.1064} \\
        \bottomrule
    \end{tabular}
    \label{tab:comparison-small-data-transposed}
\end{table}
\section{Applications of VHGM}\label{sec:applications}

VHGM is provided as a commercial web service that is accessible via a set of APIs.
As of the time of writing the paper, there are several paying customers who regularly use VHGM.

For example, a mobile phone company has a healthcare app for their phones, which encourage the users to walk more for their health.
One of the challenges of the app was how to set the appropriate goal (the number of steps the user should walk daily) because different people have different conditions. 
The VHGM has attributes on daily walking steps as well as other attributes such as the person has ``having troubles in lower back'', and the app uses this information to suggest that ``People like you but without back pain walks this number of steps daily on average'' and let the user to decide walk more.

In addition, we conducted a couple of business idea contests, asking participants for new applications based on the VHGM.
From these experiences, we observed that there are a certain ``patterns'' how the VHGM is used.

\begin{enumerate}
    \item \textbf{Estimation of a missing value from known values} -- This is the basic function of VHGM. 
    Given observed values $o_1, o_2, ..., o_m$, VHGM returns the estimated distribution $P(y|o_1, o_2, ..., o_m)$ for the target attribute $y$. 
    This pattern is useful when some attribute is hard to measure directly (e.g., measuring blood sugar usually requires an invasive process -- the VHGM provides a means to estimate the blood sugar from other observable attributes). 
    \item \textbf{What-if analysis} (Counter-factual scenario generation) -- One can provide counter-factual input to the VHGM.
    For example, ``what would my estimated BMI be if I were not smoking'' is a counter-factual query.
    These queries are useful to consider possibilities and could be useful for planning behavioral changes. 
    \item \textbf{Optimization for an desired output} -- One can use VHGM API to iteratively search possible combinations of values that would make the desired estimated value of the output attribute.
    For example, ``How can I change my diet to make the estimated risk of neuropathic pain'' would be answered by optimizing the diet attributes to make the estimated number of annual doctor visits on neuropathic pain.
    \item \textbf{Exploration of possible factors} -- One can explore possible attributes that have some relationships with the target attribute.
    For example, many senior people are concerned with their own body odor but do not know what are the possible factors that may affect body odor.
    By changing the value of the body odor attributes and see how the other 2,000+ attributes respond to the body odor attribute, one may be able to come up with hypothesis on the cause of body odor.
\end{enumerate}

This is by no means an exhaustive list.
We expect that there will be more innovative use cases of the VHGM.

%




\end{document}